\documentclass[sigconf]{acmart}

\AtBeginDocument{%
  \providecommand\BibTeX{{%
    \normalfont B\kern-0.5em{\scshape i\kern-0.25em b}\kern-0.8em\TeX}}}

\copyrightyear{2022} 
\acmYear{2022} 
\setcopyright{acmcopyright}
\acmConference[KDD '22] {Proceedings of the 28th ACM SIGKDD Conference on Knowledge Discovery and Data Mining}{August 14--18, 2022}{Washington, DC, USA.}
\acmBooktitle{Proceedings of the 28th ACM SIGKDD Conference on Knowledge Discovery and Data Mining (KDD '22), August 14--18, 2022, Washington, DC, USA}
\acmPrice{15.00}
\acmISBN{978-1-4503-9385-0/22/08}
\acmDOI{10.1145/3534678.3539359}
\usepackage{algorithmic}
\usepackage{graphicx}
\usepackage{textcomp}
\usepackage{xcolor}
\usepackage{algorithm}
\usepackage{algorithmic}
\usepackage{booktabs}
\usepackage{tabularx, makecell, multirow} 
\usepackage{adjustbox}
\usepackage{subfiles}
\usepackage{subfigure}
\usepackage{graphicx}
\usepackage{float}
\usepackage{array}
\usepackage{balance}

\newcolumntype{H}{>{\setbox0=\hbox\bgroup}c<{\egroup}@{}}
\newcommand\mymodel{KGAttack}
\newcommand{\gray}[1]{{\color{gray}~#1 }}
\begin{document}

\title{Knowledge-enhanced Black-box Attacks for Recommendations}

\settopmatter{authorsperrow=4}
\author{Jingfan Chen}
\affiliation{%
  \institution{State Key Laboratory for Novel Software Technology \\Nanjing University}
  \city{Nanjing}
  \country{China}
  }
\email{jingfan.chen@smail.nju.edu.cn}

\author{Wenqi Fan}
\affiliation{%
  \institution{The Hong Kong Polytechnic University}
  \city{Hong Kong}
  \country{China}
  }
  \email{wenqifan03@gmail.com}

\author{Guanghui Zhu}
\authornote{Corresponding author.}
\affiliation{%
  \institution{State Key Laboratory for Novel Software Technology \\Nanjing University}
  \city{Nanjing}
  \country{China}
 }
\email{zgh@nju.edu.cn}

\author{Xiangyu Zhao}
\affiliation{%
  \institution{City University of Hong Kong}
  \city{Hong Kong}
  \country{China}
  }
\email{xianzhao@cityu.edu.hk}

\author{Chunfeng Yuan}
\affiliation{%
  \institution{State Key Laboratory for Novel Software Technology \\Nanjing University}
  \city{Nanjing}
  \country{China}
  }
\email{cfyuan@nju.edu.cn}

\author{Qing Li}
\affiliation{%
  \institution{The Hong Kong Polytechnic University}
  \city{Hong Kong}
  \country{China}
  }
\email{csqli@comp.polyu.edu.hk}

\author{Yihua Huang}
\affiliation{%
  \institution{State Key Laboratory for Novel Software Technology \\Nanjing University}
  \city{Nanjing}
  \country{China}
  }
\email{yhuang@nju.edu.cn}

\renewcommand{\shortauthors}{Jingfan Chen, et al.}

\begin{abstract}

Recent studies have shown that deep neural networks-based recommender systems are vulnerable to adversarial attacks, where attackers can inject carefully crafted fake user profiles (i.e., a set of items that fake users have interacted with) into a target recommender system to achieve malicious purposes, such as promote or demote a set of target items. 
Due to the security and privacy concerns, it is more practical to perform adversarial attacks under the black-box setting, where the architecture/parameters and training data of target systems cannot be easily accessed by attackers. 
However, generating high-quality fake user profiles under black-box setting is rather challenging with limited resources to target systems. To address this challenge, in this work, we introduce a novel strategy by leveraging items' attribute information (i.e., items' knowledge graph), which can be publicly accessible and provide rich auxiliary knowledge to enhance the generation of fake user profiles.
More specifically, we propose a knowledge graph-enhanced black-box attacking framework (KGAttack) to effectively learn attacking policies through deep reinforcement learning techniques, in which  knowledge graph is seamlessly integrated into hierarchical policy networks to generate  fake user profiles for performing  adversarial black-box attacks.
Comprehensive experiments on various real-world datasets demonstrate the effectiveness of the proposed attacking framework under the black-box setting.

\end{abstract}




\maketitle



\section{Introduction}

Aiming to provide personalized services~(e.g., a list of items) to customers, recommender systems (RecSys) have been widely used in many real-world application domains~\cite{fan2022graph, fan2019deep_daso,cheng2016wide,fan2019graph,zhao2021autoloss,chen2022automated}, including social media~(e.g., Facebook, Twitter) and e-commerce~(e.g., Amazon, Taobao).
Given the powerful capacity of representation learning, Deep Neural Networks (DNNs) techniques, such as Recurrent Neural Networks~(RNNs) and Graph Neural Networks~(GNNs), have been adopted to empower recommender systems~\cite{fan2019deep,fan2020graph,zhao2021autoemb,wang2019neural,fan2018deep}.
However, recent studies have shown that most existing DNNs-based recommender systems are vulnerable to adversarial attacks~\cite{liu2021trustworthy,Fan2021JointlyAG,fang20influence,fang18poisoning,fan21attacking}, in which attackers can carefully generate fake user profiles~(i.e., a set of items that fake users have interacted with) and inject them into a target recommender system, so as to attack (i.e., promote or demote) a set of target items with malicious purposes.

\begin{figure}[t]
\centering
\centerline{\includegraphics[width=\columnwidth]{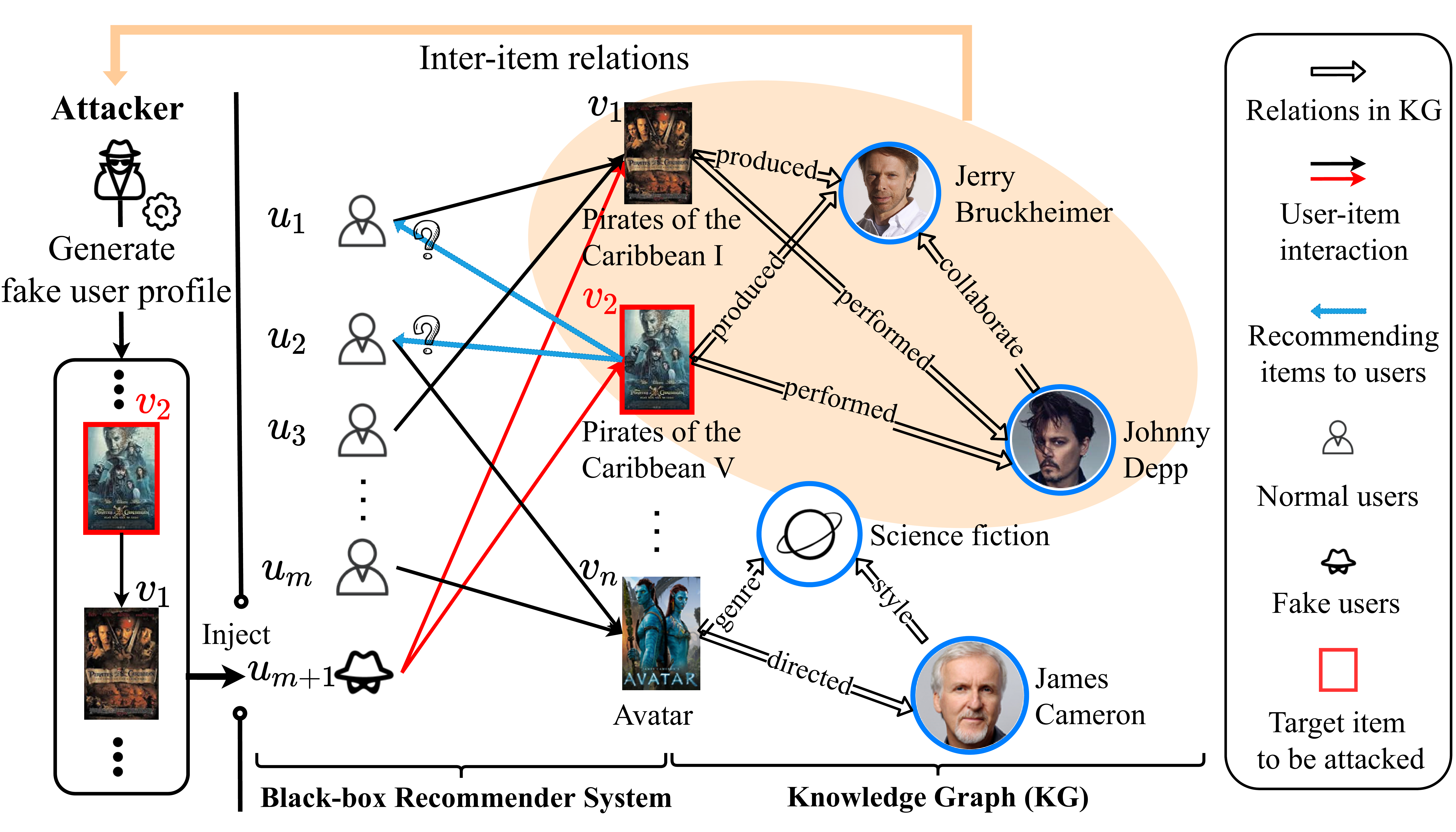}}
\vskip -0.06in
\caption{Illustration of knowledge graph-enhanced black-box attacks in recommender system. The goal of the attacker is to promote a target item $v_2$ by injecting fake user profile (i.e., user $u_{m+1} = \{..., v_{2}, v_{1} ,...\}$) based on underlying relationships (e.g., substitutes and complements) from the knowledge graph.}
\label{fig:intro}
\vskip -0.2in
\end{figure}

Due to  privacy and security concerns, compared with white-box and grey-box attacks which require full or partial knowledge of the target systems~\cite{fang20influence,li16data},  it is more practical to  perform black-box attacks  in recommender systems, which  can be achieved via query feedback to learn  generation strategies of  fake user profiles for attacking~\cite{song20poisonrec, fan21attacking}. 
Although a few works have successfully attacked  black-box  recommender systems, they are insufficient and inapplicable to generate fake user profiles for attacking recommender systems under the black-box setting. 
For example, PoisonRec~\cite{song20poisonrec} (as the very first black-box attacking method in recommender systems)  proposes to generate fake user profiles from the massive item sets by querying the target systems purely, which may make it easier to yield low-quality fake user profiles. CopyAttack~\cite{fan21attacking} proposes a copy mechanism to copy "real" user profiles from cross-domain systems under black-box setting, which is not applicable to general recommendations scenarios due to the lack of cross-domain knowledge in the real world.

To address the above problems, in this work, we propose to introduce rich auxiliary knowledge for improving the generation quality of fake user profiles. Recently, Knowledge Graph (KG) as comprehensive auxiliary data, which can introduce extra relations among items and real-world entities (from item attributes or external knowledge), have attracted increasing attention.  
In particular, KG usually contains fruitful facts and connections among items, which can be represented as a type of directed heterogeneous graph, where nodes correspond to entities~(items or item attributes) and edges correspond to relations. 
What's more, by exploring the interlinks within KG, the connectivity among items reflects their underlying relationships (e.g., substitutes and complements), which is beneficial to   integrate into the generation of fake user profiles for attacking black-box recommender systems.
Taking movies recommendations as an example  in \figurename~\ref{fig:intro}, in order to attack (i.e., promote) a target movie $v_2$ (i.e., “\emph{Pirates of the Caribbean V}”), an attacker can take advantage of target item $v_2$'s attributes (e.g.,  producers, actors, and genre), which are publicly available,  to  establish close connections between the target movie $v_2$ and  existing movies (e.g., item $v_1$"\emph{Pirates of the Caribbean I}").  
By injecting a fake user profile ($u_{m+1}$) which intentionally rates/interacts with these correlated movies into the target system, the target movie $v_2$ is most likely to be promoted/recommended to normal users (i.e., $u_1$ and $u_2$) who are interested in item $v_1$.  
To the best of our knowledge, the studies to leverage the knowledge graph for attacking black-box recommender systems  remain rarely explored. 

In this paper,  we propose a novel attacking framework (\textbf{\mymodel{}}), which employs the knowledge graph to enhance the
generation of fake user profiles from the massive item sets under the black-box setting via deep  reinforcement learning. 
More specifically, to seamlessly integrate KG into attacking strategies learning,  knowledge graph-enhanced state representation learning and knowledge-enhanced candidate selection are proposed to capture informative representations of fake user profiles  and localize   relevant item candidates, respectively. What's more, we introduce hierarchical policy networks  to generate high-quality user profiles for attacking black-box recommender systems.
The main contributions of this paper are summarized as follows:
\begin{itemize}
    \item We introduce a principle way to enhance the generation of fake user profiles by leveraging knowledge graph for attacking black-box recommender systems. To the best of our knowledge, it is the very first attempt to introduce the knowledge graph to guide fake user profiles generation;
    
    
    \item We propose a novel attacking framework~(\textbf{\mymodel{}}) under black-box setting for recommender systems, where a knowledge graph can be seamlessly integrated into hierarchical policy networks  to  effectively perform  adversarial attacks; and

    \item We conduct comprehensive experiments on various real-world datasets to demonstrate the effectiveness of the  proposed attacking method \mymodel{}.
\end{itemize}


\section{Related Work}

Recent studies have highlighted the vulnerability of DNNs-based recommender systems to adversarial attacks~\cite{huang2021data,fan21attacking,song20poisonrec}. Specifically,   attackers typically create fake user profiles to influence recommendation results of target recommender system with malicious goals.
In general, the existing attacking methods on recommender systems can be divided into three categories based on how much knowledge the attackers are able to access~\cite{fan21attacking}: white-box attacks, grey-box attacks, and black-box attacks. 
White-box and grey-box attacks suppose that the attacker has full or partial knowledge about the target recommender systems. 
For example,  works of ~\cite{fang20influence,li16data,tang2020revisiting} attempt to formulate such attacks as an optimization problem and develop gradient-based methods to learn the fake user profiles.
In addition, considering that malicious users are usually different from normal ones, generative adversarial networks~(GANs) are employed to generate malicious fake user profiles that are similar to real ones.
In \cite{chris19adversarial,lin2020attacking}, they propose GANs-based models to approximate the distribution of real users to generate fake user profiles based on estimated gradients. TripleAttack~\cite{wu21tripleattack} further extends GANs to efficiently generate fake user profiles based on TripleGAN model.

However, these white-box and grey-box  attacking methods require full or partial knowledge about the target system and training dataset, which are difficult to be carried over to the real world due to security and privacy concerns.
Without direct accessing architecture/parameters and training data of the target recommender system, black-box attacks become an emerging trend due to their security and privacy guarantees~\cite{fan21attacking,song20poisonrec,wang2018toward,yue21modelextraction}.
In~\cite{yue21modelextraction}, they extract data from the black-box recommender system and then generate fake user profiles based on the extracted data.
Recently, a few studies attempt to design fake user profiles heuristically on the black-box setting.
For instance, PoisonRec~\cite{song20poisonrec} proposes the very first deep reinforcement learning-based framework to attack black-box recommender systems.
To take advantage of cross-domain information, CopyAttack~\cite{fan21attacking} proposes a copy mechanism to copy  "real" user profiles via deep reinforcement learning techniques, instead of generating fake user profiles.
Despite the aforementioned success, they are insufficient and inapplicable to generate fake user profiles for attacking recommender
systems under the black-box setting. To address these limitations, in this paper, we propose to leverage rich auxiliary knowledge for improving the generation quality of fake user profiles. Note that items' attributes (i.e., knowledge graphs)  are rich in the real world and are publicly available~\cite{bloem2021kgbench,orlandi18Leveraging}.
To the best of our knowledge, this is the very first effort to  seamlessly integrate knowledge graphs into attacking black-box recommender systems.

\section{Problem Defination}

\textbf{Notations.} Let $U = \{u_1,...,u_m \}$ and $V = \{ v_1,...,v_n\}$ be the sets of users and items in  target recommender system, respectively. The user-item interactions  can be defined as matrix $\mathbf{Y}\in \mathbb{R}^{m\times n}$,  where $y_{ij} = 1$ indicates that user $u_i$ has interacted with item $v_j$~(e.g., clicked and purchased), and $0$ otherwise.
The user profile $P_{t} = \{v_0,...,v_{t-1}\}$ is defined as $t$ historical items a user interacted with.
In addition, in real-world application, items' external knowledge (i.e., attributes) can be accessed and formulated as knowledge graph (KG, $\mathcal{G}$) comprised of entity-relation-entity triples  $(p,r,q)$, where  $r \in \mathcal{R}$ denotes the relation, and entities $p,q\in \mathcal{V}$ denote the head and tail of a triple, respectively. For instance, a triple (\textit{Avatar, film.film.director, James Cameron}) states the fact that \emph{James Cameron} is the director of the film \emph{Avatar}. 
The goal of recommender system is to predict whether user $u_i$ will interact with an item $v_j$ via generating a top-$k$ ranked potential items $y_{i,>k} = \{v_{[1]},...,v_{[k]} \}$, where user $u_i$ is more likely to interact with $v_{[i]}$ than $v_{[i+1]}$.

\noindent \textbf{Goal of Attacking Black-box Recommender Systems}. We define the goal of attacking black-box recommender systems (\emph{promotion attacks}) is to promote a target item $v^* \in V $ by generating a set of fake users $U^{\text{F}} = \{u_{m+i}\}_{i=1}^\Delta$ with their profiles and inject them into user-item interactions $\mathbf{Y}$, where $\Delta$ is the budget given to the attacker. 
This results in the target system having a set of polluted users $U' = U \cup U^{\text{F}}$ and a polluted interaction matrix $\mathbf{Y}'\in \mathbb{R}^{(\Delta+m )\times n}$, so as to have the target item $v^*$ appear in as many users' recommendation list as possible.
Note that we consider the promotion attacks in this work only and demotion attacks as a special case of promotion attacks will be explored in the future.

 \section{The proposed Framework}

\begin{figure}[t]
\vskip -0.1in
\centering
\centerline{\includegraphics[width=\columnwidth]{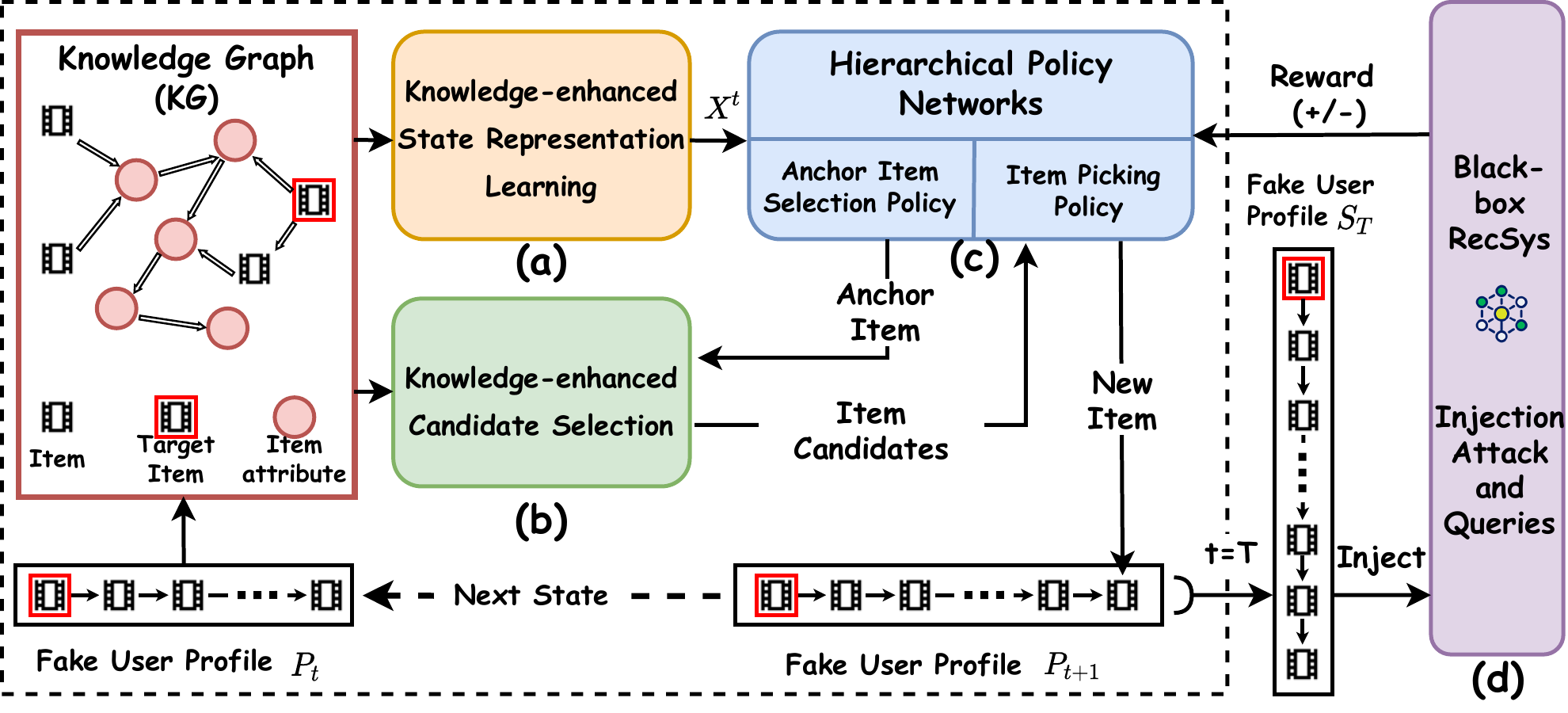}}
\vskip -0.13in
\caption{An overview of the proposal \mymodel{}. \mymodel{} first obtain the state representation based on KG (a). Then, given an anchor item,  the item candidates pool will be constructed based on both KG (b).  The anchor item is obtained by the anchor item selection policy to guide the generation of item candidates pool (c). Then, item picking policy picks a new item from the item candidates pool (c). The new item will be added to $s_t$ to obtain $s_{t+1}$. Once the length of the fake user profile reaches $T$, it will be injected into the target recommender system to  query reward for optimizing the whole framework (d).}
\label{fig:farmework}
\vskip -0.16in
\end{figure}

\subsection{An Overview of the Proposed Framework}


The goal of this work is to conduct black-box attacks on a target recommender system.
To achieve the goal, we propose a knowledge-enhanced attacking framework (\textbf{\mymodel{}}), which adopts a deep reinforcement learning to learn attacking strategies of  generating fake user profiles via query feedback. 
The overview framework of the proposed method \mymodel{} is shown in Figure~\ref{fig:farmework}, which consists of four main components:  knowledge-enhanced state representation learning (a), knowledge-enhanced candidate selection (b), hierarchical policy networks (c), and injection attack and queries (d).
Knowledge-enhanced state representation learning  aims to learn the representation of fake user profiles  via taking advantage of KG at the current step.
The component of knowledge-enhanced candidate selection is introduced to leverage the KG to generate  item candidates pools of the anchor item.
In order to effectively generate fake user profiles in large-scale discrete action space (i.e., items) and enable efficient exploration, we propose hierarchical policy networks to decompose such actions into two steps: (1) localizing an anchor item (Anchor Item Selection Policy), and (2) picking the next item from the item candidates pool (Item Picking Policy).  
The last  component  aims to perform injection attacks and query the black-box systems to have the reward for updating the whole framework. 
Next, we will first detail the attacking environment of the proposed black-box reinforcement learning method.

\noindent \textbf{Attacking Environment.}
In this paper, we model the proposed black-box attacks procedure  as a Markov Decision Process (MDP), 
in which an attacker~(agent) interacts with environment~(i.e., the target recommender system) by sequentially generating fake user profiles and injecting them into the environment, so as to maximize the expected cumulative reward from the environment.  Formally, the MDP is a tuple with five elements  $(\mathcal{S}, \mathcal{A}, \mathcal{P}, \mathcal{R}, \gamma)$ as:
\begin{itemize}
\item \emph{State space} $\mathcal{S}$:  
The state $s_t \in \mathcal{S}$ is defined as a chronologically sorted sequence of items a fake user interacted~(i.e., fake user profile) before time $t$. In general, the state can be encoded as representation of fake user profile $\mathbf{x}_t$. 

\item \emph{Action space $\mathcal{A}$}: The action $a_t = (a^\text{anchor}_t, a^\text{item}_t) \in \mathcal{A}_t$ is to determine two actions.  The first action $a^\text{anchor}_t$ is to determine a specific anchor item $v_t^{\text{anchor}}$  for generating the item candidates pool $\mathcal{C}_{t}$, and another one $a^\text{item}_t$ is to pick an item  from $\mathcal{C}_{t}$ for generating  fake user profiles.


\item \emph{Transition probability} $\mathcal{P}$: Transition probability $p(s_{t+1} |s_t, a_t)$ is defined as the probability of state transition from the current $s_t$ to  next state $s_{t+1}$ when  attacker takes action $a_t$.

\item  \emph{Reward} $\mathcal{R}$: After the attacker injects a fake user profile at state $s_t$, the target recommender system provides immediate feedback/reward $r_t$ (i.e., Top-$k$ recommended items on spy users) via query. The detailed definition can be found in section~\ref{sec:injection}. 



\item  \emph{Discount factor} $\gamma$: Discount factor $\gamma \in [0,1]$ is to measure the importance of future rewards. $\gamma = 0$ will make the agent only consider current rewards, and $\gamma = 1$ will make the agent strive for a long-term high reward.
\end{itemize}

\noindent \textbf{RL Attacking Policy.}
The goal of attacker is to seek an optimal policy $\pi:\mathcal{S} \rightarrow \mathcal{A}$ which can maximize the expected cumulative reward as follows:
\begin{align}
    \max _{\pi} \mathbb{E}_{\tau \sim \pi}[R(\tau)], \text { where } R(\tau)=\sum_{t=0}^{|\tau|} \gamma^{t} r\left(s_{t}, a_{t}\right)
\end{align}
where $\tau = (s_0,a_0,...,s_{T-1},a_{T-1})$ denotes the $T$-length trajectory~(multiple state-action pairs) that are generated based on the policy.
To achieve the goal, we utilize the Deep Reinforcement Learning (DRL)~\cite{lillicrap2015continuous} with neural networks to automatically learn the attack policy.
$\mathbb{E}_{\pi_{\theta}}[\sum^{T-1}_{l=t} \gamma^{l-t} r_l]$ from any state~(i.e., expected value function $V^{\pi_{\theta}}(s_t)$), where $\theta$ denotes the policy parameters and $\gamma$ denotes the discount factor.
To tackle the challenges of the large-scale discrete action space (i.e., a huge number of items) and sparse rewards, we build the attacking policy based on the Advantage Actor-Critic~(A2C) architecture, consisting of an actor network and a critic network~\cite{mnih16asy,schulman2017proximal}.
The actor network $\pi_{\theta}(a_t|s_t)$~(also called  policy network) parameterized with $\theta$ generates the distribution over actions $a_t$ based on the state $s_t$ and its possible action space $\mathcal{A}_t$. The actor network is updated based on the policy gradient. Meanwhile, the critic network $V_{\omega}(s_t)$ parameterized with $\omega$ is employed to accurately estimate the contribution of each action to the rewards for better guiding the actor in the sparse reward scenario. In other words, the critic network is used to judge whether the selected action matches the current state. 
What's more, to reduce the variance and have better convergence properties, we calculate the advantage value $A_{\omega}(s_t)$ as the actual critic judgment based on the output of the critic network~\cite{mnih16asy}.
The critic network is updated according to the temporal difference~(TD) approach~\cite{silver16mastering} by minimizing the following squared error as follows:
\begin{equation}\label{tderror}
    \mathcal{L_{\omega}} = \sum_t \left(\sum_{j=0}^{T-t}\gamma^{j} r_{t+j}-V_{\omega}(s_t)  \right)^2
\end{equation}
Finally, based on the judgment from the critic network, the actor network optimizes its’ parameters to maximize the attacking performance so as to output better actions in the next iteration.

\subsection{Knowledge-enhanced State Representation Learning}

The detail of knowledge-enhanced state representation learning is illustrated in Figure~\ref{fig:KGAttack_components} (a). 
With a target item $v^*$ to be attacked (i.e., promoted), the current fake user profile at step $t$ can be denoted as $P_{t} = \{v^*,v_1,...,v_{t-1} \}$. The goal of this component is to encode  a profile $P_{t}$ and treat the learned  representation  $\mathbf{x}_{t}$ as current state $s_t$.

\subsubsection{\textbf{Knowledge-enhanced Item Representation Initialization}}
As items can be linked with entities in KG, the semantic and correlation information among items can provide informative signals to enhance item representation learning.  Therefore, it is desirable to improve  representations of  fake user profiles by taking advantage of the knowledge graph.
To achieve the goal, an auxiliary task in KG is introduced to initialize item (i.e., entities) representations, where TransE~\cite{bordes2013translating}, as one of the most representative techniques for knowledge embedding techniques, is used to pre-train the representation of entities and relations in KG.
More specifically,  the entities' representations are optimized by minimizing a margin-based criterion as follows:
 \begin{equation}
 \mathcal{L}_\text{pre-train} = \sum_{(p,r,q) \in \mathcal{B}^+}\sum_{(p',r,q')\in \mathcal{B}^-} 
 [ d(\mathbf{p}+\mathbf{r},\mathbf{q}) + \xi - d(\mathbf{p'}+\mathbf{r},\mathbf{q'}) ]_+
 \end{equation}
where $[x]_+ = \text{max}(0,x)$. $\xi$ denotes margin between positive and negative triples.
$\mathcal{B}^+$ includes positive triples satisfying $\mathbf{p}+\mathbf{r} \simeq \mathbf{q} $ and  $\mathcal{B}^-$ includes negative triples having $\mathbf{p}+\mathbf{r} \neq \mathbf{q}$. $d$ can be either $L_1$ or $L_2$ norm. $\mathbf{p}$, $\mathbf{q}$, and $\mathbf{r}$ are the    embeddings of entity/item or relation. 
For simplicity, the initial knowledge-enhanced item representation can be denoted as  $\mathbf{e}^0_{i}$ for each interacted item $v_i$ in the fake user profile $P_t$ at step $t$.

\subsubsection{\textbf{Knowledge-enhanced Item Representation}}
With item representations initialized via an auxiliary knowledge embedding task, we further propose to utilize Graph Neural Networks (GNNs) to  learn  item representations with knowledge graph. 
The intuition of using GNNs is that nodes' representation can be naturally captured via node feature and topological structure on the graph under deep neural networks paradigm~\cite{kipf2017semi}. 
More specifically, item representations can be enhanced via local neighbors (e.g., items or item attributes)  aggregations in KG. Moreover,  a relation attention $\alpha$ is proposed to characterize different relationships between entities in heterogeneous KG. Mathematically, a knowledge-enhanced representation of an interacted item $v_i$ can be defined as follows:
\begin{align}
     \mathbf{e}^{l}_i &= \mathbf{W}^{l}_1  \cdot \mathbf{e}^{l-1}_i + \mathbf{W}^{l}_2 \cdot \sum_{v_j\in \mathcal{N}(v_i)}\alpha^{l}_{i,j}  \mathbf{e}_j^{l-1},\\
         \alpha^{l}_{i,j} &= \text{softmax}\Bigg(  \big( \mathbf{W}_{\text{in}}  \cdot \mathbf{e}^{l-1}_i \big)^{\top} \big(  \mathbf{W}_{\text{out}}  \cdot \mathbf{e}^{l-1}_j   \big) /\sqrt{d}  \Bigg)
\end{align}
where $\mathbf{e}^{l}_i$ denotes the embedding of an interacted item $v_i$ in the fake user profile at layer $l$, and $\mathcal{N}(v_i)$ denotes the local neighbor entities~(e.g., item or item attributes) of an item $v_i$. $d$ denotes the neighbor size of item $v_i$. $\mathbf{W}^l_1$ and $\mathbf{W}^l_2$ are  trainable weights at layer $l$. $\mathbf{W}_\text{in}$ and $\mathbf{W}_\text{out}$ are trainable weights shared by each layer.

\begin{figure}[t]
\vskip -0.06in
\centering
\centerline{\includegraphics[width=0.99\columnwidth]{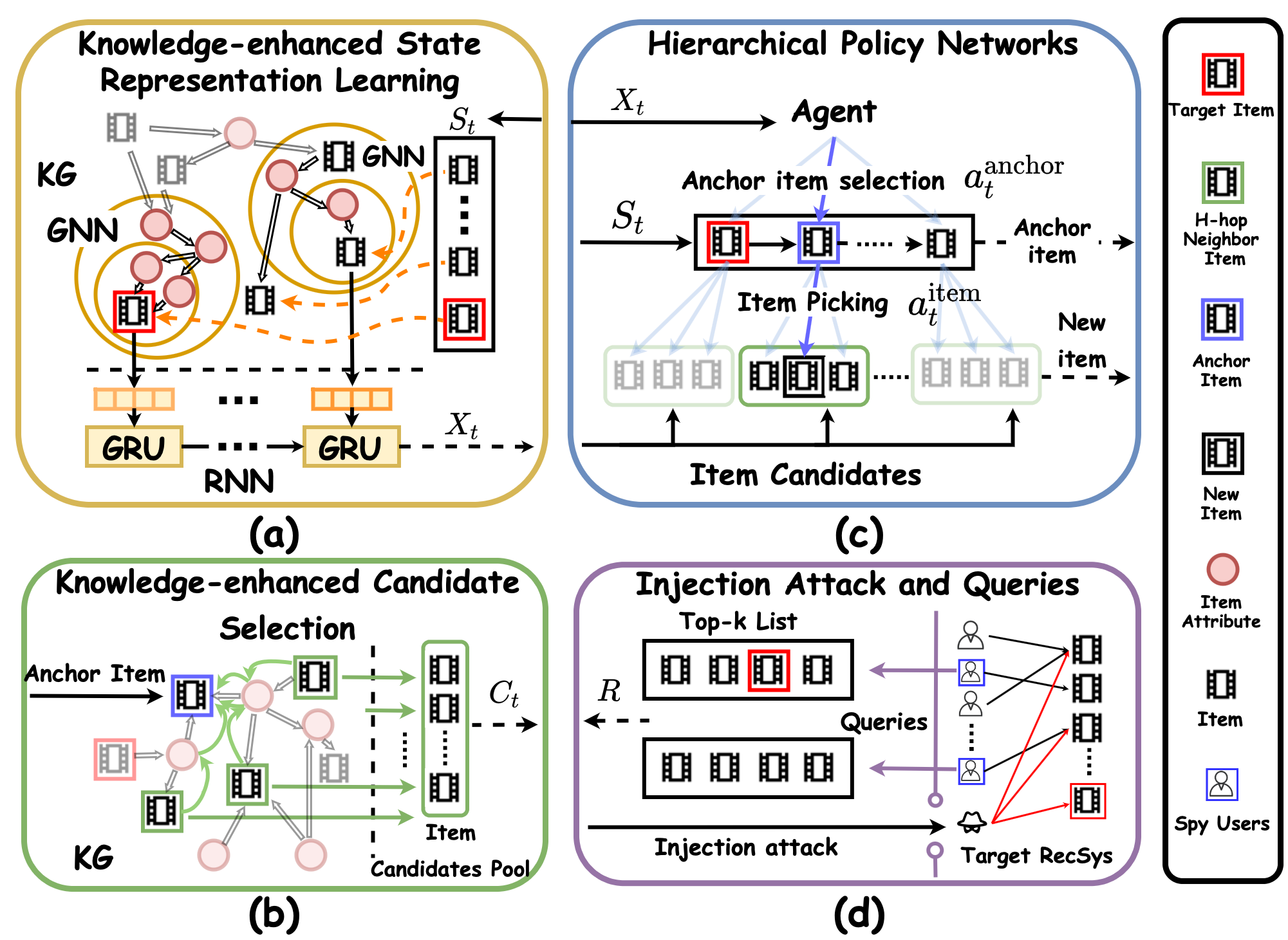}}
\vskip -0.1in
\caption{A detailed illustration of (a) knowledge-enhanced state representation learning, (b) knowledge-enhanced candidates selection, (c) hierarchical policy networks, and (d) injection attack and queries.}
\label{fig:KGAttack_components}
\vskip -0.16in
\end{figure}

\subsubsection{\textbf{State Representation Learning}}
With knowledge-enhanced item representation $\mathbf{e}_{i}^l$, we propose to learn the state representation ($\mathbf{x}_{t}$) on $P_{t}$ at state $t$ for policy network. Since the fake user profile contains sequential dependency among items~(i.e., long and short term), we introduce to employ a Recurrent Neural Network (RNN) to encode  fake user profiles. Specifically, we adopt an RNN with a gated recurrent unit (GRU) as the network cell~\cite{cho2014learning} to capture fake user's historical behaviors $P_{t} = \{v^*,v_1,...,v_{t-1} \}$ based on the knowledge-enhanced item representation $\mathbf{e}_{i}^l$ as follows:
\begin{equation}
\begin{aligned}
&\mathbf{z}_{t}=\sigma_r\left(\mathbf{W}_{z} \mathbf{e}_{t}^l+\mathbf{U}_{z} \mathbf{h}_{t-1}+\mathbf{b}_{z}\right) \\
&\mathbf{u}_{t}=\sigma_u\left(\mathbf{W}_{u}\mathbf{e}_{t}^l+\mathbf{U}_{u} \mathbf{h}_{t-1}+\mathbf{b}_{u}\right) \\
&\mathbf{\hat{h}}_{t}=\sigma_c \left(\mathbf{W}_c \mathbf{e}_{t}^l+\mathbf{U}_c\left(\mathbf{z}_{t} \circ \mathbf{h}_{t-1}\right)+\mathbf{b}_c\right) \\
&\mathbf{h}_{t}=\mathbf{u}_{t} \circ \mathbf{h}_{t-1}+\left(1-\mathbf{u}_{t}\right) \circ \mathbf{\hat{h}}_{t} \\
\end{aligned}
\end{equation}
where $\mathbf{z}_t$ and $\mathbf{u}_t$ denote the reset gate and update gate vector respectively. $\circ$ is the Hadamard product. $\mathbf{W}_*$, $\mathbf{U}_*$ and $\mathbf{b}_*$ denote the trainable parameters. Then, the obtained hidden vector $\mathbf{h}_{t}$ is set as state representation $\mathbf{x}_{t}$ at state $t$. As a result, the state representation contains rich auxiliary knowledge toward the target item $v^*$ from  item attributes.

\subsection{Knowledge-enhanced Candidate Selection}
\label{sec: Action candidates selection}

To efficiently learn attacking strategies based on the policy network, a key challenge is how to handle the sample efficiency problem, which is caused by the large action space consisting of a large number of candidate items (i.e., actions)~\cite{zou2019reinforcement}. To this end, as shown in \figurename~\ref{fig:KGAttack_components} (b), we propose  knowledge-enhanced candidate selection to employ KG for localizing some relevant item candidates (i.e., \emph{item candidates pool} $\mathcal{C}_{t}$) which  share similar attributes (e.g., substitutes and complements). 
Specifically, given an  anchor item $v_t^{\text{anchor}}$ (please see section~\ref{sec:hpn} for more details) at step $t$,  we generate an \emph{item candidates pool} $\mathcal{C}_{t}$ by extracting the $H$-hop neighbors of the anchor item $v_t^{\text{anchor}}$ in KG.
To achieve the goal, we first construct the $H$-hop relevant entity set in KG ($\mathcal{G}$) as follows:
\begin{equation}
\begin{split}
    \mathcal{E}_{t}^h = \{q|(p, r, q) \in \mathcal{G}, p\in \mathcal{E}_{t}^{h-1}  \}, 
    h=1, 2, ..., H,
\end{split}
\end{equation}
where $\mathcal{E}_{t}^0 = v_t^{\text{anchor}}$. 
Then, to construct the item candidates pool $\mathcal{C}_{t}$, we collect a fixed size  items candidates in  $H$-hop relevant entity set as follows:
\begin{equation}
\label{eq:candidates}
\mathcal{C}_{t} = \{ v|v\in \bigcup_{h=1}^H \mathcal{E}_{t}^{h}, v \in V  \}
\end{equation}

\subsection{Hierarchical Policy Networks}
\label{sec:hpn}

With the knowledge-enhanced state representation $\mathbf{x}_{t}$, the hierarchical policy networks aim to learn attacking strategies for the generation of fake user profiles sequentially.
What's more, to effectively generate fake user profiles
in large-scale discrete action space (i.e., items) and enable efficient
exploration, we decompose attacking strategies into hierarchical actions: anchor item selection $a_t^{\text{anchor}}$ and item picking $a_t^{\text{item}}$, as shown in \figurename~\ref{fig:KGAttack_components} (c). 
An anchor item $v_t^{\text{anchor}}$ is first selected from the current fake user profile $P_t$ via  $a_t^{\text{anchor}}$. 
Then, a new item $v_{t}$ can be generated from the item candidates pool $\mathcal{C}_{t}$ of an anchor item $v_t^{\text{anchor}}$ via  $a_t^{\text{item}}$, which will be added to the current fake user profile.
To achieve hierarchical actions, we employ two actor networks: anchor item policy $\pi_{\theta}^{\text{anchor}}(a_t^{\text{anchor}}|s_t)$  and item picking  policy $\pi_{\phi}^{\text{item}}(a_t^{\text{item}}|s_t)$ parameterized with $\theta$ and $\phi$, respectively. 

\noindent \textbf{Anchor Item Selection.} Note that the size of $a_t^{\text{anchor}}$'s action space for anchor item  at step $t$ can vary dynamically. To achieve such dynamic action space scenario, in practice, we adopt mask mechanism to mask parts of fixed output vector (i.e., $T$-length) on policy $\pi_{\theta}^{\text{anchor}}(a_t^{\text{anchor}}|s_t)$ to avoid sampling invalid positions. Formally, an anchor item $v_t^{\text{anchor}}$ is sampled via shallow neural networks with state representation $\mathbf{x}_{t}$  as input as follows:
\begin{equation}
    \pi^{\text{anchor}}_{\theta}(a_t^{\text{anchor}}|s_t) = \text{Softmax}(\mathbf{W}_{\text{A},2}\text{ReLU}(\mathbf{W}_{\text{A},1}\mathbf{x}_{t})+\mathbf{m}_t)
\end{equation}
where $\theta=\{\mathbf{W}_{\text{A},1}, \mathbf{W}_{\text{A},2}\}$ are the learned parameters. $\mathbf{m}_t$ denotes the $T$-length masked vector to ignore the invalid action positions due to dynamic action space scenario.

\noindent \textbf{Item Picking.} With the selected anchor item $v_t^{\text{anchor}}$, we can generate a fixed size item candidate pool $\mathcal{C}_{t}$ for item picking action.  
As illustrated in the second level of the hierarchy in Figure~\ref{fig:KGAttack_components} (c), we model the $\pi_{\phi}^{\text{item}}$ by calculating the similarity score between the state representation $\mathbf{x}_{t}$ and each item in $\mathcal{C}_{t}$:
\begin{align}
    \mathbf{\hat{x}}_t = \text{ReLU}&(\mathbf{W}_{\text{I},1}\mathbf{x}_{t})\\
    \pi_{\phi}^{\text{item}}(a_t^{\text{item}}|s_t) =& \frac{\text{exp}(\mathbf{W}_{\text{I},2}[\mathbf{\hat{x}}_{t};\mathbf{e}_t])}{\sum_{v_j\in\mathcal{C}_{t}}\text{exp}(\mathbf{W}_{\text{I},2} [\mathbf{\hat{x}}_{t};\mathbf{e}_j])}
\end{align}
where $\phi=\{\mathbf{W}_{\text{I},1},\mathbf{W}_{\text{I},2}\}$ are trainable parameters. $\mathbf{e}_j$ denotes the knowledge-enhanced representation of item $v_j$ in item candidate pool. $[\cdot ; \cdot]$ is the concatenation operator. 

\noindent \textbf{Exploration-exploitation Trade-off.} To fully take advantage of the target item's knowledge, a greedy strategy is to set the target item as an anchor item and pick the target item's neighbors to generate fake user profiles. The main limitation of this strategy lies in its lack of exploration: if the target item's attribute knowledge is not informative enough,  the generation of fake user profiles can be `stuck'  and keep choosing a sub-optimal decision forever without any chances to discover better options (i.e., items). Therefore, to achieve an exploration-exploitation trade-off, we introduce an anchor ratio $\epsilon$ to find a balance between the target item exploitation and the policy networks exploration for the generation of fake user profiles.
More specifically, at each step $t$, attacker selects an anchor item from $P_{t}$ with $\epsilon$ probability based on policy network $\pi_{\theta}^{\text{anchor}}(a_t^{\text{anchor}}|s_t)$ , while setting the target item $v^*$ as the anchor item with $1-\epsilon$ probability.

\subsection{Injection Attack and Queries}
\label{sec:injection}
The last component aims to inject the generated fake user profiles $P_{t}$ into the target recommender system. 
To access the system and get feedback for updating policies under the black-box setting, inspired by CopyAttack~\cite{fan21attacking}, a set of spy users was established  in the target recommender system for performing query and receiving the reward to optimize the framework  after injection. 
More specifically, Top-$k$ recommended items on spy users are the feedback/rewards upon querying the target recommender systems as follows:
\begin{align}\label{reward}
    r_t = 
    \begin{cases}
    \frac{1}{|\hat{U}|}\sum^{|\hat{U}|}_{i=1} \text{HR}(\hat{u}_i, v^*, k),&t=T-1;\\
    0& t=0,...,T-2,
    \end{cases}
\end{align}
where $\text{HR}(\hat{u}_i, v^*, k)$ returns the hit ratio for target item $v^*$ in the top-$k$ list of the spy user $\hat{u}_i \in \hat{U}$.
If  target item $v^*$ appears in the spy user's recommendation list,  HR is set to $1$, otherwise $0$.
Note that the set of spy users $\hat{U}$ can be a subset of normal users who are already established in the target recommender system before the attacks. 
In practice, when the length of generated user profile $P_{t}$ reaches pre-defined $T$, we perform injection and queries.
Next, we introduce the details of the \mymodel{} training.

\subsection{Model Training}

Considering the sparse reward and expensive cost of interacting with the target recommender system, the proximal policy optimization algorithm~(PPO) in DRL~\cite{schulman2017proximal} is adopted to train the proposed \mymodel{} for attacking black-box recommender systems. Each training iteration mainly includes two stages: trajectory generation and model parameters updating. The whole training process is detailed in Appendix~\ref{sec:algorithm}.
In the first stage, the agent (attacker) simultaneously generates $N$ fake user profiles by repeatedly performing the hierarchical policy networks. 
Then, these $T$-length fake user profiles will be injected into the target recommender system. The reward $r_t$ is calculated based on Equation~(\ref{reward}). In practice, the obtained single reward is shared for $N$ trajectories.
Finally, the agent stores the transition $(s_t, a^\text{anchor}_t,a^{\text{item}}_t, r_t, s_{t+1})$ into the replay memory buffer $\mathcal{D}$.
In the second stage, the two actor network and  critic network are updated according to the transitions in the replay memory buffer $\mathcal{D}$.
The actor network $\pi_{\theta}^{\text{anchor}}(a_t^{\text{anchor}}|s_t)$ is updated by maximizing the following surrogate clipped objective~\cite{schulman2017proximal}:
\begin{equation}
\label{policy gradient}
\begin{split}
\theta_{\text{New}} = \arg \max_{\theta} \frac{1}{NT}\sum_{i=1}^N \sum_{t=0}^{T-1} \min 
\Big(\frac{\pi_{\theta}(a_t|s_t)}{\pi_{\theta_{\text{Old}}}(a_t|s_t)}A_{\omega}(s_t),\\
\text{clip}( \frac{\pi_{\theta}(a_t|s_t)}{\pi_{\theta_{\text{Old}}}(a_t|s_t)},1-\psi,1+\psi )A_{\omega}(s_t)
\Big)
\end{split}
\end{equation}
where $\theta_{\text{Old}}$ denotes the old parameters of the actor network. $\psi$ is the hyperparameter used to clip the probability ratio.
The item picking policy network's parameter $\phi$ is updated in the same way. 
The critic network is updated via TD approach according to Equation~(\ref{tderror}).

\section{Experiments}

In this section, we conduct extensive experiments to demonstrate the effectiveness of the proposed framework.

\subsection{Experimental Settings}
\subsubsection{\textbf{Datasets}}

We utilize the following three real-world datasets to verify the effectiveness of our proposed  \mymodel{}: 

\textbf{MovieLens-1M~(ML-1M)} contains user-item interactions for movies recommendations on the MovieLens website. \textbf{Book-Crossing} records interactions between users and books in the Book-Crossing community. \textbf{Last.FM} is   musician listening datasets from Last.FM online music system.
The knowledge graphs of three datasets are extracted from Microsoft Satori and are released by~\cite{wang2018ripplenet,wang19knowledge}. The statistics of the datasets are detailed in Appendix~\ref{sec:dataset statistics}.

\subsubsection{\textbf{Attacking Environment: Target Recommender System}}

\begin{itemize}
\item \textbf{Evasion Attack} (model testing stage): at this stage, target model's parameters are frozen without any retraining on polluted dataset. To achieve evasion attack, we adopt inductive GNNs-based recommendation  \textbf{Pinsage}~\cite{ying18graph} as target recommender system to be attacked, which aims to learn the user and item representations based on the user-item graph via local propagation mechanism.

\item \textbf{Poison Attack} (model training stage): at this stage, the training data can be poisoned/changed once injecting  fake user profiles. Then, the target recommender is required to retrain on the poisoned dataset. 
We adopt a general deep recommendation method \textbf{NeuMF}~\cite{he2017neural} as the target model for poison attack.
What's more, to leverage item's knowledge graph, we also conduct poison attacks on a target system \textbf{KGCN}~\cite{wang19knowledge} (knowledge-enhanced GNNs based recommendation method).

\end{itemize}
The implementation details (e.g., hyper-parameters of \mymodel{} and the target recommender systems) are listed in Appendix~\ref{sec:implementation details}.

\begin{table*}[htbp]
\vskip -0.12in
  \centering
  \caption{Performance comparison of different black-box attacks method on target recommender system  Pinsage. We use bold fonts and underline to label the best performance and the best baseline performance, respectively. H@$k$ and N@$k$ denote HR@$k$ and NDCG@$k$, respectively.}
    \vskip -0.12in
        \scalebox{0.8}
{
    \begin{tabular}{c|c|c|c|c|c|c|c|c|c|c|c|c}
    \toprule
    \toprule
    \multirow{2}[4]{*}{Dataset} & \multicolumn{4}{c|}{MovieLens-1M (ML-1M)} & \multicolumn{4}{c|}{Book-Crossing} & \multicolumn{4}{c}{Last.FM} \\
\cmidrule{2-13}          & \multicolumn{1}{c}{H@20} & \multicolumn{1}{c}{H@10} & \multicolumn{1}{c}{N@20} & N@10  & \multicolumn{1}{c}{H@20} & \multicolumn{1}{c}{H@10} & \multicolumn{1}{c}{N@20} & N@10  & \multicolumn{1}{c}{H@20} & \multicolumn{1}{c}{H@10} & \multicolumn{1}{c}{N@20} & N@10 \\
    \midrule
    Without Attack & 0.000  & 0.000  & 0.000  & 0.000  & 0.191  & 0.095  & 0.065  & 0.042  & 0.193  & 0.012  & 0.073  & 0.005  \\
    RandomAttack & 0.000  & 0.000  & 0.000  & 0.000  & 0.202  & 0.092  & 0.069  & 0.041  & 0.152  & 0.092  & 0.054  & 0.040  \\
    TargetAttack & 0.464  & 0.056  & 0.118  & 0.017  & 0.706 & 0.370  & 0.226  & 0.141  & 0.242  & 0.042  & 0.064  & 0.014  \\
    TargetAttack-KG & 0.398  & 0.028  & 0.099  & 0.008  & 0.862 & 0.606  & 0.342  & 0.276  & 0.282  & 0.110  & 0.085  & 0.043  \\
    \midrule
    PoisonRec & 0.610  & \underline{0.138}  & 0.162  & \underline{0.047}  & 0.930 & 0.748  & \underline{0.428}  & \underline{0.381}  & \underline{0.442}  & \underline{0.148}  & \underline{0.125}  & \underline{0.052}  \\
    PoisonRec-KG & \underline{0.628}  & 0.108  & \underline{0.163}  & 0.035  & \underline{0.930} &\underline{0.748}  & 0.427  & 0.380  & 0.438  & 0.148  & 0.123  & 0.051  \\
    \midrule
    KGAttack-Target & 0.554  & 0.009  & 0.144  & 0.029  & \textbf{0.940} & 0.780 & 0.437 & 0.396 & 0.442  & 0.144  & 0.125  & 0.051 \\
    KGAttack-Seq & 0.504  & 0.009  & 0.132  & 0.031  & 0.932 & 0.750  & 0.425  & 0.379  & 0.436  & 0.148  & 0.123  & 0.051  \\
    KGAttack & \textbf{0.672 } & \textbf{0.184 } & \textbf{0.183 } & \textbf{0.063 } & 0.934 & \textbf{0.788}  & \textbf{0.459}  & \textbf{0.422}  & \textbf{0.452 } & \textbf{0.152 } & \textbf{0.130 } & \textbf{0.053}  \\
    \bottomrule
    \bottomrule
    \end{tabular}%
    }
  \label{tab:attack pinsage}%
\end{table*}%

\subsubsection{\textbf{Baselines}}
Though there are various attacking methods~\cite{fang20influence,li16data} developed for the white/grey-box setting, they cannot be easily used in  black-box setting. Therefore, we select the following seven baselines:
(1) \textbf{RandomAttack}: This baseline randomly samples items from the entire item space $V$ to construct the fake user profiles.
(2) \textbf{TargetAttack}: This baseline incorporates the target item $v^*$ in the fake user profiles and randomly samples the remaining items from the entire item space $V$.
(3) \textbf{TargetAttack-KG}: This baseline is similar to TargetAttack, while remaining items are randomly sampled from the $H$-hop neighbors of the target item.
(4) \textbf{PoisonRec~\cite{song20poisonrec}}: This is a state-of-the-art black-box attacking method, which adopt deep reinforcement learning to generate fake user profiles.  
(5) \textbf{PoisonRec-KG}: This method is a variant of PoisonRec. Here, we adopt knowledge graph to enhance item representations learning via GNNs and constrain the RL action search space. 
(6) \textbf{KGAttack-Target}: This method is a variant of our proposed model, which only considers the target item as the anchor item.
(7) \textbf{KGAttack-Seq}: This method is a variant of our model, which sets the last interacted item in each fake user profile as the anchor item.

\subsection{Overall Attacking Performance Comparison}

\subsubsection{\textbf{Evasion Attack (Pinsage)}.}
We first compare the attacking performance of different attacking methods on \textbf{Pinsage} recommender system under the black-box setting, as shown in Table~\ref{tab:attack pinsage}. 
We have the following main findings:
\begin{itemize}
\item As a naive black-box attacking method, RandomAttack does not gain any remarkable improvement compared with \emph{WithoutAttack}. By contrast, the attacking performance  can be significantly improved in black-box recommender systems when generating fake user profiles with the target item (i.e., TargetAttack).

\item  For random sampling-based attacking method, TargetAttack-KG  performs better than TargetAttack on Book-Crossing and Last.FM datasets, which can be attributed to  rich auxiliary knowledge (i.e., knowledge graph) from  items. 
However, we also can observe that TargetAttack-KG cannot perform better than TargetAttack on ML-1M dataset, which might be attributed to  a large number of entities in ML-1M's knowledge graph.

\item  DRL-based attacking methods (e.g., PoisonRec and KGAttack) perform better than random sampling-based attacking methods (e.g., RandomAttack and TargetAttack), which indicates that the DRL-based method can learn  effective  strategies to perform  black-box attacks via query feedback in recommender systems.

\item  \mymodel{}-Target can perform better than \mymodel{}-Seq, which implies that the  greedy strategy  to set the target item as an anchor item can exploit the target item’s neighbors (i.e., knowledge graph) to generate fake user profiles. 
What's more, in most cases, our proposed method KGAttack  can achieve the best attacking performance in black-box recommender systems under different evaluation metrics, except KGAttack-Target on HR@20 metric, which suggests that our proposed  anchor item selection strategy with anchor ratio  $\epsilon$  can  achieve exploration-exploitation trade-off in most cases.
These promising attacking performance on KGAttack also supports that the proposed hierarchical policy networks via anchor item selection and item picking can effectively dig out the relevant items of the target item via knowledge graph to enhance  fake user profiles generation.

\end{itemize}

\subsubsection{\textbf{Poison Attack (KGCN and NeuMF)}}
\begin{figure}[t]
\centering

\subfigure[KGCN: HR@20]{\label{fig:reset_hr}\includegraphics[width=0.48\columnwidth]{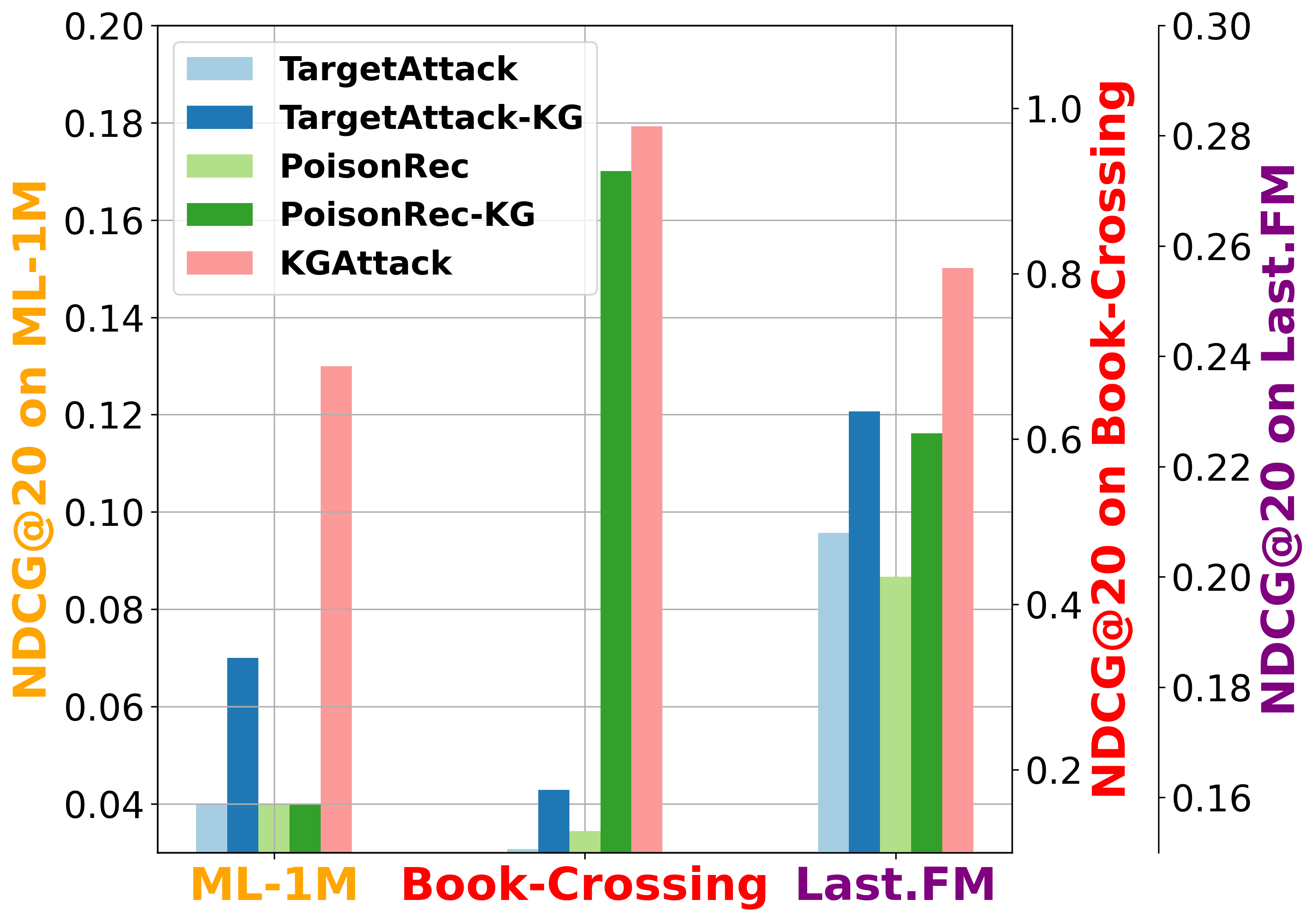}}
\subfigure[KGCN: NDCG@20]{\label{fig:reset_ndcg}\includegraphics[width=0.48\columnwidth]{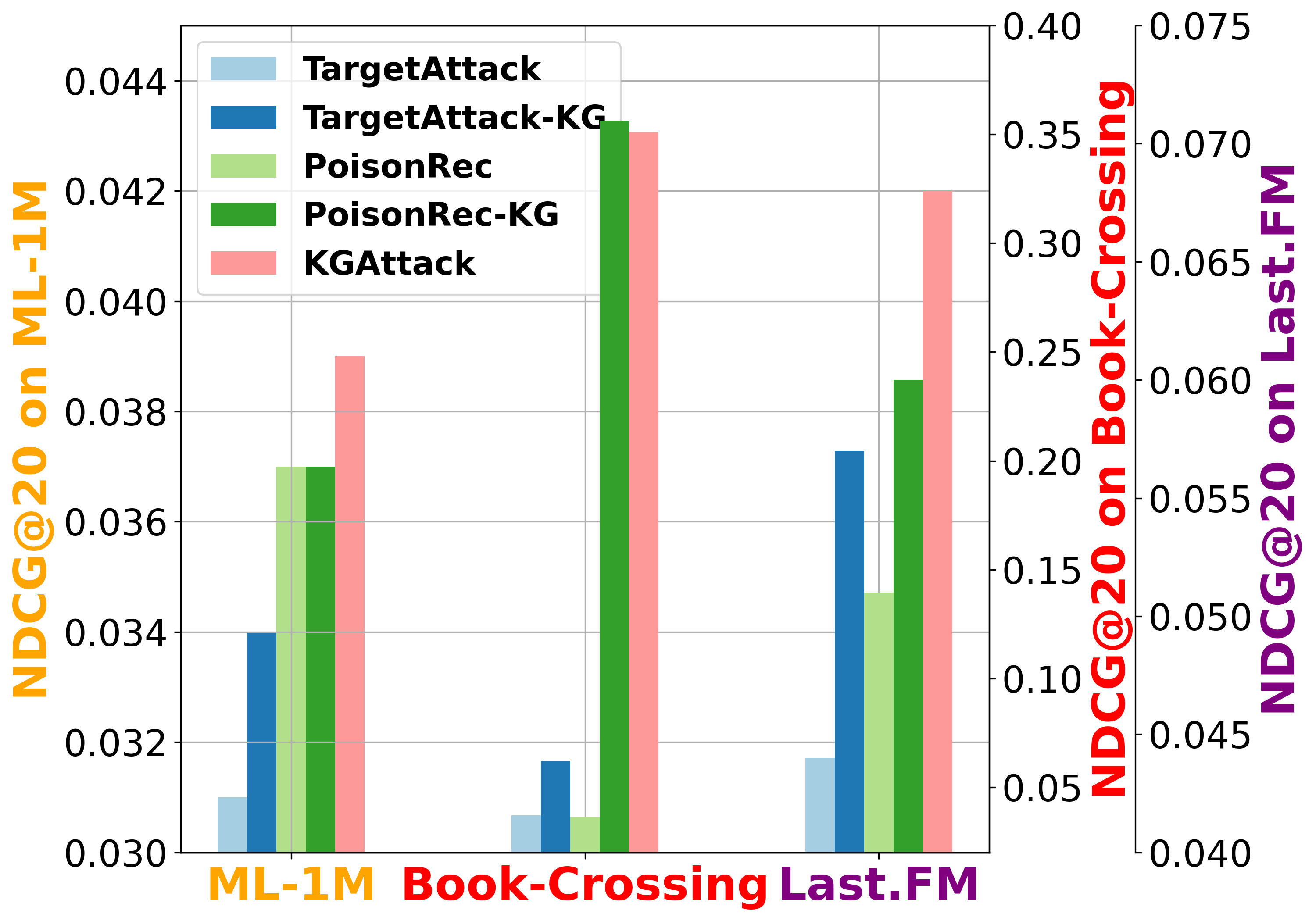}}
\subfigure[NeuMF: HR@20]{\label{fig:neumf_hr20}\includegraphics[width=0.48\columnwidth]{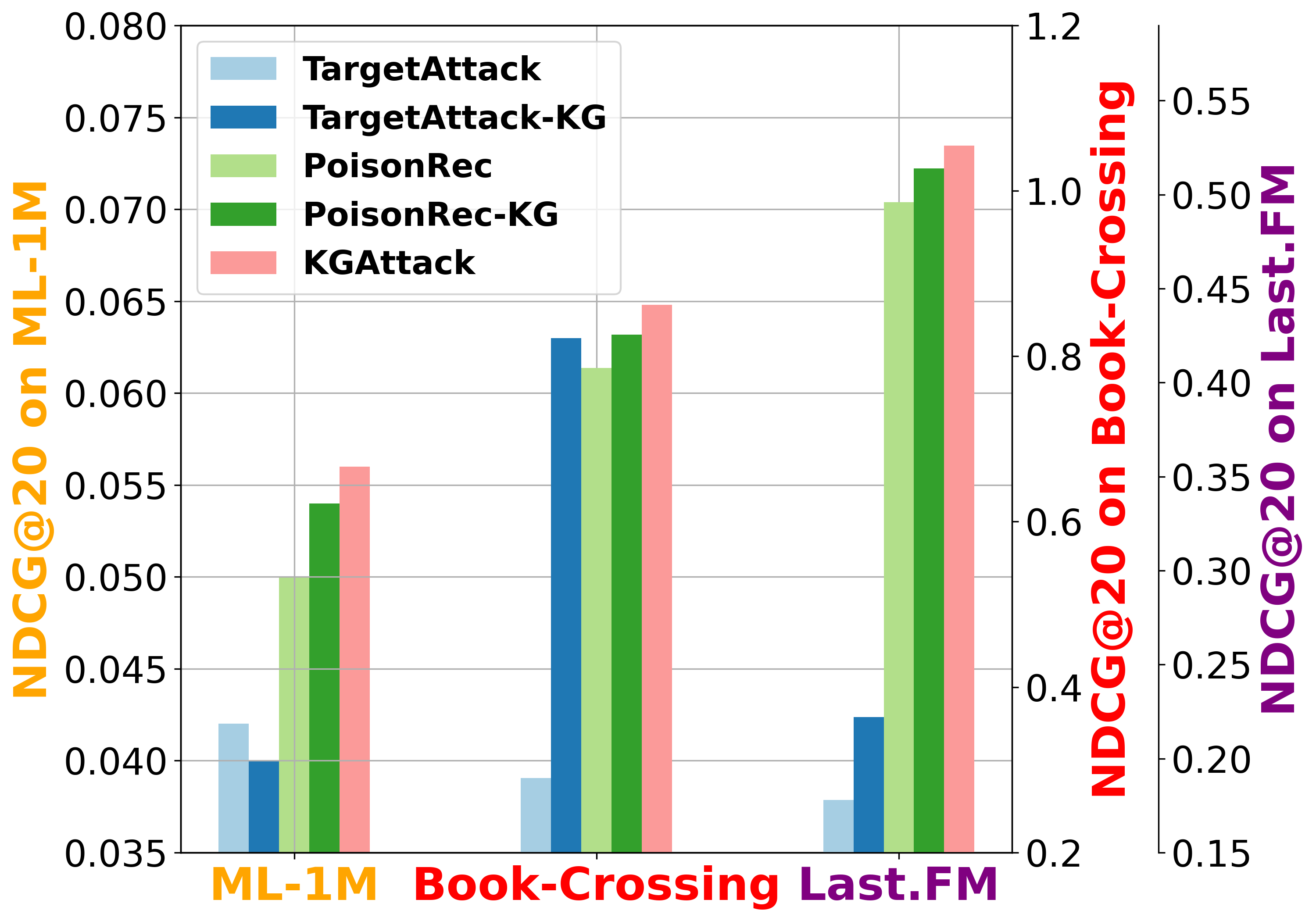}}
\subfigure[NeuMF: NDCG@20]{\label{fig:reset_ndcg}\includegraphics[width=0.48\columnwidth]{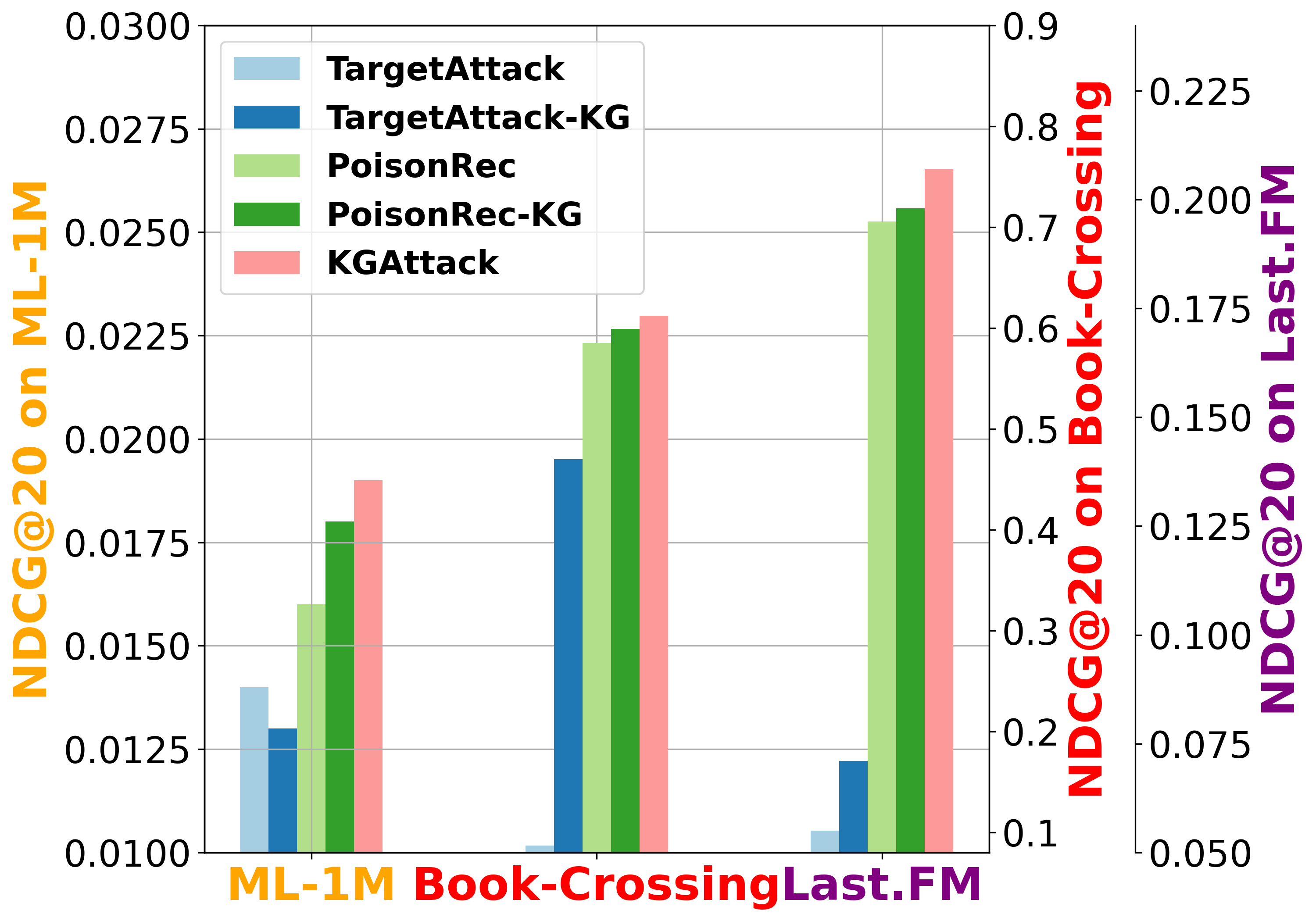}}
\caption{Performance comparison of different attacking methods across two metrics HR$@20$ and NDCG$@20$ on KGCN and NeuMF.}

\label{fig:performance-neumf}
\vskip -0.2in
\end{figure}

We evaluate the attacking performance under different black-box recommender systems (i.e.,  \textbf{KGCN} and \textbf{NeuMF}) on three datasets. The experimental results are shown in Figure~\ref{fig:performance-neumf}. Due to the space limitation, we only present the results on two metrics HR@20 and NDCG@20. We do not show the results on some baselines (e.g., Without Attack, RandomAttack, KGAttack-Target, and KGAttack-Seq) since similar observations as Table~\ref{tab:attack pinsage} can be made. 
We can observe that the models' performance behaves differently on two target models. KG-incorporated methods~(e.g., TargetAttack-KG and PoisonRec-KG) significantly improve the attacking performance on KGCN. 
This suggests that  leveraging items' knowledge graph can help  attack black-box recommender systems. This is due to the fact that such knowledge-enhanced recommender systems fully take advantage of  connectivity among item to improve recommendation performance.
What's more, these experimental results indicate that the seamless integration between KG and hierarchical policy networks is beneficial to generate high-quality fake user profiles for attacking black-box systems.

In addition, we observe that our proposed \mymodel{} improves the attacking performance on NeuMF under the black-box setting. 

Our proposed \mymodel{} can almost beat all baselines on these two target models, which proves the effectiveness of the hierarchical policy networks with knowledge-enhanced for the fake user profiles generation. In addition, these results also demonstrate that our proposed attacking model is applicable to  attack most types of recommender systems.

\begin{table}[b]
  \vskip -0.15in
  \centering
  \caption{Comparison between \mymodel{} and its variants for Pinsage. Bold fonts denotes the best performance.}
    \scalebox{0.8}
{
  \begin{adjustbox}{max width=\columnwidth,center}
    \begin{tabular}{c|c|c|c|c|c|c}
    \toprule
    \toprule
    \multirow{2}[4]{*}{Models} & \multicolumn{2}{c|}{MoveLens-1M} & \multicolumn{2}{c|}{Book-Crossing} & \multicolumn{2}{c}{Last.FM} \\
\cmidrule{2-7}          & \multicolumn{1}{c}{H@20} & \multicolumn{1}{c|}{N@20} & \multicolumn{1}{c}{H@20} & \multicolumn{1}{c|}{N@20} & \multicolumn{1}{c}{H@20} & \multicolumn{1}{c}{N@20} \\
    \midrule
    KGAttack (-KGE) &    0.598   &    0.163   &   0.928    &   0.442    &  0.422     & 0.119 \\
    KGAttack (-GNN) &   0.630    &    0.161   &   0.926    &   0.442    &    0.446   & 0.124 \\
    KGAttack (-Relevant) &  0.628 & 0.163 &  0.930 & 0.427&    0.438   & 0.123  \\
    KGAttack (-HPN) &   0.532    &    0.140   &   0.926    &   0.421    &    0.430   & 0.121 \\
    \midrule
    KGAttack &  \textbf{0.672} &  \textbf{0.183}  &  \textbf{0.934}   &   \textbf{0.459}    &   \textbf{0.460}    &  \textbf{0.130} \\
    \bottomrule
    \bottomrule
    \end{tabular}%
    \end{adjustbox}
    }
  \label{tab:ablation}%
\end{table}%

\subsection{Ablation Study}
In this section, we conduct the ablation studies to study the effectiveness of different components in  \mymodel{}. 
We compare our model with its four variants: 
(1) \textbf{\mymodel{} (-KGE)}: this method removes the knowledge-enhanced item representation initialization (i.e., knowledge graph embedding), and utilizes the randomly initialized item embedding. 
(2) \textbf{\mymodel{} (-GNN)}: this method directly utilizes RNN to provide the state representation, which only models the temporal representation of the fake user profile without considering the knowledge-enhanced item representation.
(3) \textbf{\mymodel{} (-Relevant)}: this method removes the knowledge-enhanced candidate selection, which utilizes  all items in KG as the action space for item picking.
(4) \textbf{\mymodel{} (-HPN)}: this method removes the hierarchical policy networks, which directly picks items from large-scale discrete action space (i.e., items) to generate the fake user profiles without  anchor item selection.

The comparison results are shown in Table~\ref{tab:ablation}.
Comparing either \mymodel{}~(-KGE) or \mymodel{}~(-GNN) with   \mymodel{}, the attacking performance improvement indicates that the knowledge-enhanced items representations enable the fake user profile's representation learning which encodes   items' semantic correlation information. 
The comparison between \mymodel{} (-Relevant) and \mymodel{} shows that  leveraging the knowledge-enhanced candidate selection can efficiently reduce the action space and further boost the attacking performance.
In addition, the performance difference between \mymodel{} (-HPN) and \mymodel{} validates that the proposed anchor item selection in hierarchical policy networks is effective to guide the selection of the item candidates, which can further enhance the generation of fake user profiles.

\subsection{Parameter Analysis}
This subsection studies the impact of the model hyper-parameters.

\subsubsection{\textbf{Effect of anchor ratio}}

\begin{table}[htb]
  \centering
  \caption{Effect of anchor ratio with HR@20 for Pinsage.}
    \scalebox{0.8}
{
    \begin{tabular}{c|ccccc}
    \toprule
    \toprule
    $\epsilon$   & 0.1   & 0.3   & 0.5   & 0.7   & 0.9 \\
    \midrule
    MovieLens-1M & 0.582  & 0.534  & 0.620  & 0.622  & \textbf{0.660 } \\
    Book-Crossing & 0.916  & 0.920  & \textbf{0.934 } & 0.928  & 0.930  \\
    Last.FM & 0.432  & 0.444  & 0.442  & \textbf{0.460 } & 0.448  \\
    \bottomrule
    \bottomrule
    \end{tabular}%
    }
  \label{tab:para anchor ratio}%
\end{table}%

In this subsection, we analyze the effect of anchor ratio $\epsilon$. Note that  anchor ratio $\epsilon$ is proposed to achieve an exploration-exploitation trade-off.
The experimental results are given in Table~\ref{tab:para anchor ratio}.
We can see that attacking performance can achieve better when the value of the anchor ratio is larger than $0.5$.
This indicates that our proposed \mymodel{} prefers  selecting  anchor item via hierarchical policy networks (i.e., anchor item selection  $a_t^{\text{anchor}}$). 
We also observe that encouraging the target item as the anchor item excessively will degrade the attacking performance. Therefore, for anchor item selection, we need to  carefully find a balance between policy networks exploration and the target item exploitation for the generation of fake user profiles.

\subsubsection{\textbf{Effect of hop number}}

\begin{table}[htb]
\vskip -0.15in
  \centering
  \caption{Effect of hop number for constructing item candidates pool with HR@20 for Pinsage.}
  \scalebox{0.8}
{
    \begin{tabular}{c|cccc}
    \toprule
    \toprule
    $H$     & 1     & 2     & 3     & 4 \\
    \midrule
    MovieLens-1M & 0.648 & \textbf{0.672}  & 0.514  & 0.608  \\
    Book-Crossing & 0.926  & 0.934  & \textbf{0.940 } & 0.938  \\
    Last.FM & 0.450  & 0.452  & \textbf{0.460 } & 0.452  \\
    \bottomrule
    \bottomrule
    \end{tabular}%
}
  \label{tab:para hop}%
\end{table}%

We vary the hop number $H$ for knowledge-enhanced item candidate selection in \mymodel{} to further investigate the changes of attacking performance. 
The experimental results are shown in Table~\ref{tab:para hop}.
We can observe that  \mymodel{} can  achieve the best performance when $H=3$ on Book-Crossing and Last.FM datasets, while  $H=2$ on ML-1M dataset. 
We attribute the observation to the relative richness of connectivity information among items via knowledge graph (e.g., substitutes and complements): the KG on ML-1M dataset contains more rich correlations information than that of Book-Crossing and Last.FM datasets.
Thus, too large  $H$ on ML-1M dataset  may  bring many noises during item picking action from the item candidates pool.

\subsubsection{\textbf{Effect of budget}}
\begin{figure}[htb]
\subfigure{\label{fig:budget_ml_hr}\includegraphics[width=0.87\linewidth]{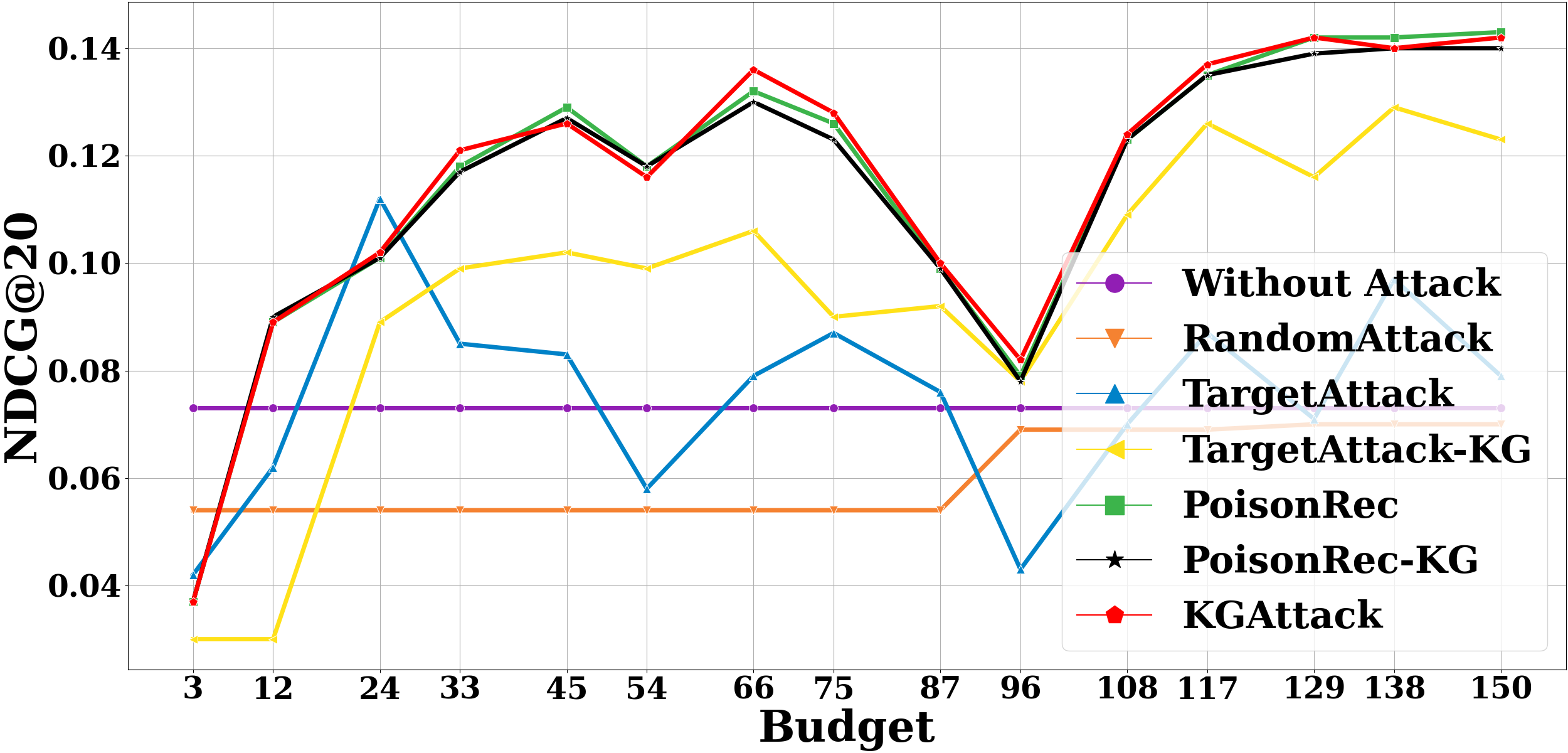}}
\vskip -0.15in
\caption{Effect of budget~(the number of injected fake user profile) on Last.FM over NDCG@20 for Pinsage.}
\label{fig:budget lastfm}
\vskip -0.15in
\end{figure}

Here, we investigate the black-box attacking performance by varying the budget $\Delta$ (i.e., the number of fake user profiles). Due to the space limitation, we show the results for Pinsage on Last.FM over NDCG@20.
The experimental results are shown in Figure~\ref{fig:budget lastfm}.  In general,  we can observe that RL-based attack methods (e.g., PoisonRec and KGAttack) can gain better performance, since these methods can adjust their attacking policy to maximize the expected cumulative long-term reward. 
In addition, when the attacking budget is larger than 30, KGAttack can achieve the best performance in most times, which demonstrates the effectiveness of our proposed framework.

\section{Conclusion}

In this work, we propose a knowledge-enhanced attacking framework for black-box recommender systems (\textbf{\mymodel{}}), which can leverage knowledge graph to enhance the generation of fake user profiles from the massive item sets under the black-box setting via deep reinforcement learning. What's more, our framework provides advanced components to perform state representation learning and learn attacking strategies via hierarchical policy networks, so as to generate high-quality fake user profiles. Extensive experiments on three real-world datasets demonstrate the effectiveness of our proposed attacking method under the black-box setting. What's more, we conducted comprehensive experiments on different target recommender systems, such as deep recommendation method (NeuMF), GNNs based recommender system (Pinsage), and knowledge-enhanced GNNs based recommendation method (KGCN). 
In addition,  the ablation study also demonstrates the effectiveness of the designed components in our proposed attacking framework in recommender systems.

\begin{acks}
Jingfan Chen, Guanghui Zhu, Chunfeng Yuan, and Yihua Huang are supported in part by the National Natural Science Foundation of China (NSFC) (No.62102177 and No.U1811461), the Natural Science Foundation of Jiangsu Province (No.BK20210181), the Key R\&D Program of Jiangsu Province (No.BE2021729), Open Research Projects of Zhejiang Lab (No.2022PG0AB07), and the Collaborative Innovation Center of Novel Software Technology and Industrialization, Jiangsu, China. 
Wenqi Fan and Qing Li are supported by NSFC (No.62102335) and a General Research Fund from the Hong Kong Research Grants Council (Project No.: PolyU 15200021). 
Xiangyu Zhao is supported by the APRC - CityU New Research Initiatives (No.9610565, the Start-up Grant for the New Faculty of the City University of Hong Kong), the SIRG - CityU Strategic Interdisciplinary Research Grant (No.7020046, No.7020074), and the CCF-Tencent Open Fund. 
\end{acks}


\bibliographystyle{ACM-Reference-Format}
\balance
\bibliography{references/reference}

\newpage
\clearpage

\appendix

\section{Appendix}

\subsection{Statistics of Datasets}\label{sec:dataset statistics}
The statistics of the three datasets are summarized in Table~\ref{tab:datasets}. 

\begin{table}[b]
  \vskip -0.15in

  \centering
  \caption{Basic statistics of three datasets. "\# Items in KG" indicates the number of items that appeared in both the knowledge graph and the dataset. "Avg. $H$-hop NBR" denotes the average number of the $H$-hop neighbors for each item.}
  \vskip -0.1in
  \resizebox{\columnwidth}{!}{
    \begin{tabular}{c|c|c|c|c}
    \toprule
    \toprule
          &   Attribute    & MovieLens-1M & Book-Crossing & Last.FM \\
    \midrule
    \multirow{4}[2]{*}{Dataset} & \# Users & 5,950 & 13,097 & 1,874 \\
          & \# Items & 3,532 & 306,776 & 17,612 \\
          & \# Interactions & 574,619 & 1,149,772 & 92,780 \\
          & \# Items in KG & 2,253 & 14,114 & 3,844 \\
    \midrule
    \multirow{6}[2]{*}{KG} & \# Entities & 182,011 & 77,903 & 9,366 \\
          & \# Relations & 12    & 25    & 60 \\
          & \# KG triples & 1,241,995 & 151,500 & 15,518 \\
          & Avg. 1-hop NBR & 27    & 15    & 5 \\
          & Avg. 2-hop NBR & 298   & 24    & 14 \\
          & Avg. 3-hop NBR & 1,597 & 82    & 60 \\
    \bottomrule
    \bottomrule
    \end{tabular}%
    }
  \label{tab:datasets}%
\vskip -0.20in
\end{table}%

\subsection{Implementation Details}\label{sec:implementation details}
 Our proposed framework is implemented on the basis of PyTorch. Following the common practice~\cite{sun20are}, to split the dataset, we leverage each user's last $30\%$ interacted items as a testing set, while the remaining are left for training.
To evaluate the quality of the target recommender system, we adopt two widely-used ranking metrics: Hit Ratio~(HR@$k$) and Normalized Discounted Cumulative Gain~(NDCG@$k$) by following the previous work~\cite{he2017neural,fan21attacking}, where $k$ is set to  $20$ and $10$.
The first metric measures whether a test item is appeared in the top-$k$ recommendation list, while the latter measures the ranking positions of the test items. 
We randomly sample $10$ target items with less than $10$ interactions for promotion attacks. To observe the attacking performance on two metrics, we randomly sample $50$ spy users and $500$ normal users who did not interact with the target items from testing set for \mymodel{}'s training reward and evaluation, respectively. Following~\cite{fan21attacking}, for each  user, we construct the test candidates set  by randomly sampling $100$ items that  users did not interact with. Note that the target items do not interact with these users either.
Then, we attack each target item and observe whether it appeared in the spy users' top-$k$ list among these $100$ test candidates after attacks.
To obtain the feedback (reward) from the target recommender system, we query spy users per $N=3$ injections. The budget $\Delta$ is set to $75$.
We set the $H=2$ for constructing  item candidates pool.  The size of the item candidates pool is set to 50 for Book-Crossing and Last.FM, and 200 for ML-1M according to their average item neighbors.

Note that we use the same target recommender systems on all baselines for fair comparison. Meanwhile,  we use the 'light'/shallow target recommender systems to evaluate the final performance due to the efficiency. 
The hyper-parameters of  \mymodel{} and different   target recommender systems   are detailed as below:
\begin{itemize}
 \item \textbf{Pinsage}: Training epochs are set to 1000. The layers are set to 2. The first layer dimension is 64 for ML-1M, 32 for Book-Crossing, and 16 for Last.FM. The second layer dimension is 16 for ML-1M and Book-Crossing, and 8 for Last.FM. For attack setting, we set the fake profile length $T=16$ for ML-1M and Book-Crossing, and 4 for Last.FM.
 
 \item \textbf{NeuMF}: Training epochs are set to 10 for ML-1M and Last.FM, and 20 for Book-Crossing. We set the layers of NeuMF as 2 for ML-1M, 4 for Book-Crossing, and 1 for Last.FM. The layer dimension is set to 32 for all layers. For attack setting, we set the fake user profile length $T=16$ for ML-1M and Book-Crossing, and 4 for Last.FM. 
 
\item \textbf{KGCN}: Training epochs are set to 10 for ML-1M, and 5 for both Book-Crossing and Last.FM. We set the layers of KGCN as 2. The layer dimension is set to 16 for all layers and all datasets. For attack setting, we set the fake user profile length $T=16$ for ML-1M and Book-Crossing, and 8 for Last.FM.
\end{itemize}

\subsection{Training Process of \mymodel{}}\label{sec:algorithm}

We summarize the training process of the proposed \mymodel{} in Algorithm 1.
\begin{algorithm}[htbp]
\label{algo}
\caption{\textbf{KGAttack}}
\begin{algorithmic}[1]
\STATE Randomly initialize the Actor $\pi_{\theta}$, $\pi_{\phi}$ and Critic $V_{\omega}$ with parameters $\theta$, $\phi$ and $\omega$.
\STATE Initialize replay memory buffer $\mathcal{D}$
\FOR{episode number $c$ in $[0,\Delta / N)$}
\STATE \gray{// \textbf{(i) Trajectory Generation}}
\STATE \FOR{fake user $i$ in $[m+cN+1,m+(c+1)N+1]$}
\STATE Initialize state $s_0$ based on $P_{0,u_i}=\{v_*\}$
\FOR{step $t$ in $[0,T-1]$}
\STATE Select anchor item $v_t^{\text{anchor}}$ according to $\pi^{\text{anchor}}_{\theta}$ with anchor ratio $\epsilon$
\STATE generate the item candidates $\mathcal{C}_{t,u_i}$ according to $v_t^{\text{anchor}}$
\STATE Pick a new item $v_t^{\text{}}$ according to $\pi^{\text{item}}_{\theta}$ and $\mathcal{C}_{t,u_i}$
\STATE Obtain state $s_{t+1} = \{s_t,v_t^{\text{}}\}$ and reward $r_t$ 
\STATE Push $\{s_t,a_t^{\text{item}},a_t^{\text{anchor}},r_t,s_{t+1}\}$ into the memory buffer $\mathcal{D}$
\ENDFOR
\ENDFOR
\STATE 
\STATE \gray{//\textbf{(ii)  Networks Update}}
\STATE Get transitions from replay memory buffer $\mathcal{D}$
\STATE Update the critic network $V_{\omega}$ by minimizing the loss in Equation~(\ref{tderror}) 
\STATE Update the actor networks $\pi_{\theta},\pi_{\phi}$ by maximizing Equation~(\ref{policy gradient}) via stochastic gradient ascent with Adam.
\STATE Clean replay memory buffer $\mathcal{D}$
\ENDFOR

\end{algorithmic}
\end{algorithm}


\end{document}